\documentclass[sn-mathphys-num]{sn-jnl}

\usepackage{graphicx}%
\usepackage{multirow}%
\usepackage{amsmath,amssymb,amsfonts}%
\usepackage{amsthm}%
\usepackage{mathrsfs}%
\usepackage[title]{appendix}%
\usepackage{xcolor}%
\usepackage{textcomp}%
\usepackage{manyfoot}%
\usepackage{booktabs}%
\usepackage{algorithm}%
\usepackage{algorithmicx}%
\usepackage{algpseudocode}%
\usepackage{listings}%
\usepackage{hyperref}%
\usepackage{tabularx}%
\usepackage{booktabs}%
\usepackage{longtable}%
\usepackage{adjustbox}%
\usepackage{float}%
\usepackage{rotating}%
\usepackage{placeins}%
\usepackage{csquotes}

\renewcommand{\thetable}{\Roman{table}}

\theoremstyle{thmstyleone}%
\theoremstyle{thmstyletwo}%

\theoremstyle{thmstylethree}%

\begin{document}

\title[Systematic mapping]{Rebalancing the Scales: A Systematic Mapping Study of Generative Adversarial Networks (GANs) in Addressing Data Imbalance}


\author*[1,1]{\fnm{Pankaj} \sur{Yadav}}\email{yadav.58@iitj.ac.in}

\author[2,2]{\fnm{Gulshan} \sur{Sihag}}\email{sihag.1@alumni.iitj.ac.in}

\author[3,1]{\fnm{Vivek} \sur{Vijay}}\email{vivek@iitj.ac.in}

\affil*[1]{\orgdiv{Department Of Mathematics}, \orgname{Indian Institute of Technology Jodhpur}, \orgaddress{\street{Karwar}, \city{Jodhpur}, \postcode{342037}, \state{Rajasthan}, \country{India}}}

\affil[2]{\orgdiv{Department of Computer Science}, \orgname{UPHF, CNRS, UMR}, \orgaddress{\street{LAMIH}, \city{Valenciennes}, \postcode{8201}, \state{Hauts-de-France}, \country{France}}}


\abstract{Machine learning algorithms are used in diverse domains, many of which face significant challenges due to data imbalance. Studies have explored various approaches to address the issue, like data preprocessing, cost-sensitive learning, and ensemble methods. Generative Adversarial Networks (GANs) showed immense potential as a data preprocessing technique that generates good quality synthetic data. This study employs a systematic mapping methodology to analyze 3041 papers on GAN-based sampling techniques for imbalanced data sourced from four digital libraries. A filtering process identified 100 key studies spanning domains such as healthcare, finance, and cybersecurity.\\

Through comprehensive quantitative analysis, this research introduces three categorization mappings as application domains, GAN techniques, and GAN variants used to handle the imbalanced nature of the data. GAN-based oversampling emerges as an effective preprocessing method. Advanced architectures and tailored frameworks helped GANs to improve further in the case of data imbalance.    GAN variants like vanilla GAN, CTGAN, and CGAN show great adaptability in structured imbalanced data cases. Interest in GANs for imbalanced data has grown tremendously, touching a peak in recent years, with journals and conferences playing crucial roles in transmitting foundational theories and practical applications.\\

While with these advances, none of the reviewed studies explicitly explore hybridized GAN frameworks with diffusion models or reinforcement learning techniques. This gap leads to a future research idea develop innovative approaches for effectively handling data imbalance.
}




\maketitle

\section{Introduction}\label{sec1}


The application of machine learning (ML) covers several domains and has surged in recent years due to its ability to automate complex tasks. However, imbalanced data (a scenario where the distribution of classes is skewed) poses challenges for ML algorithms~\cite{1},~\cite{2},~\cite{3}. The imbalanced nature of data can be accessed through a ratio called imbalanced ratio (IR). When the IR is low, minority classes are often treated as noise, leading to biased models and reduced accuracy~\cite{4},~\cite{5}. This imbalance can result in higher false positives (FP) and fewer true positives (TP), hindering the accuracy of ML applications~\cite{6}. Many solutions are discovered to address these challenges and are typically categorized into data-level preprocessing~\cite{7},~\cite{8},~\cite{9},~\cite{10} and algorithmic adjustments~\cite{11},~\cite{12},~\cite{13}. While algorithmic approaches include cost-sensitive learning~\cite{14},~\cite{15} and ensemble methods~\cite{16},~\cite{17}. These methods focus on optimizing ML algorithms for imbalance scenarios, but they often lack reusability across datasets. Data-level techniques, on the other hand, address imbalance through methods like undersampling~\cite{18},~\cite{19},~\cite{20},~\cite{21}, oversampling~\cite{22},~\cite{23},~\cite{24},~\cite{25},~\cite{26}, or hybrid sampling~\cite{27},~\cite{28},~\cite{29},~\cite{30}.

\begin{figure}
    \centering
    \includegraphics[height=0.65\textheight,width=1.0\textwidth]{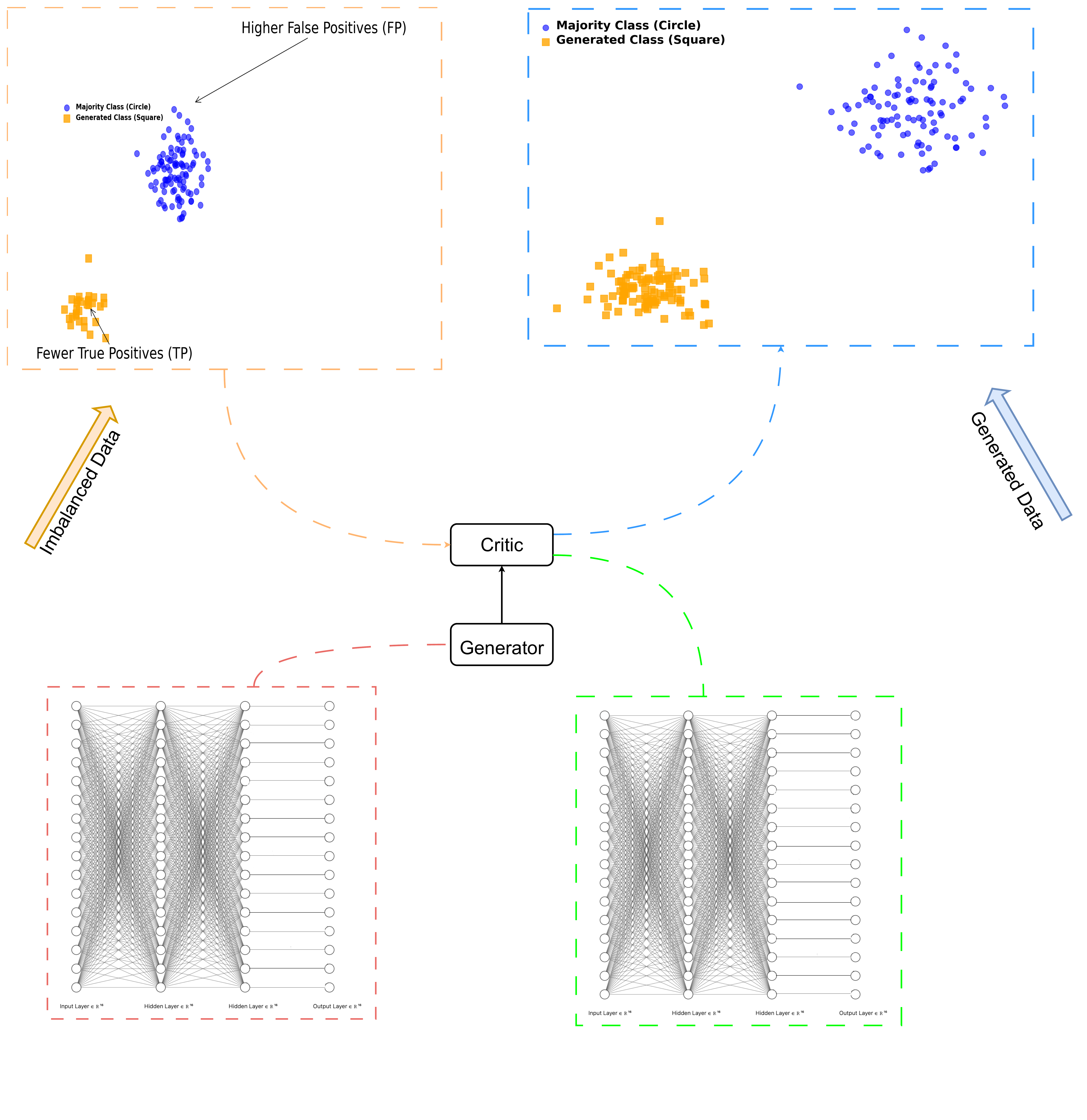}
    \caption{An overview of the workflow of imbalanced data and GANs}
    \label{fig:work}
\end{figure}

Generative Adversarial Networks (GANs)~\cite{31},~\cite{32},~\cite{33},~\cite{34},~\cite{35} have emerged as a powerful data-level solution for addressing imbalanced datasets. GANs use adversarial learning to generate high quality synthetic samples by augmenting minority classes~\cite{36},~\cite{37}. Their adaptability shows innovation across diverse domains extending to structured data~\cite{38},~\cite{39},~\cite{40},~\cite{41}. Despite their potential, systematic studies exploring the use of GANs for data imbalance remain sparse, resulting in gaps in understanding their domain-specific applications, architecture variations, and integration with hybrid techniques. The workflow illustrating the handling of imbalanced data using GANs~\cite{42},~\cite{43},~\cite{44},~\cite{45} is presented in Fig.~\ref{fig:work}. The figure provides a detailed presentation of the process, starting with identifying an imbalanced dataset, followed by applying a GAN-based framework. The framework generates synthetic data points through a learned neural network called generator~\cite{46},~\cite{47}. Simultaneously, a critic or discriminator~\cite{48},~\cite{49},~\cite{50},~\cite{51} evaluate these instances, distinguishing between real and synthetic instances. This iterative process improves the generator's ability to synthesize real data points. Ultimately, this results in a balanced dataset with an improved representation of minority class.

The study aims to systematically map and analyze the role of GANs in handling data imbalance. The key contributions of this paper are as follows: 
\begin{itemize}
    \item A comprehensive review of 3041 studies from four digital libraries, refined through a rigorous ten-step filtering process to identify 100 key works.
    \item Development of three categorization mappings, including application domains, different GAN techniques, and various GAN variants, offering a structured overview of GAN-based approaches in addressing imbalanced datasets.
    \item In-depth analysis of advanced data-level strategies and performance metrics employed with GANs in handling imbalanced data.
    \item Insights into the growing interest of GANs in addressing data imbalance, supported by theoretical and practical advancements in recent years.
\end{itemize}

\section{Related Work}
Several reviews and surveys have explored the role of GANs in addressing data imbalance with a focus on specific application domains~\cite{52},~\cite{53},~\cite{54},~\cite{55}, improved methodologies~\cite{56},~\cite{57},~\cite{58}, and architecture adaptations~\cite{59},~\cite{60},~\cite{61},~\cite{62}. The research methodology identified four major reviews and surveys~\cite{63},~\cite{64},~\cite{65},~\cite{66}. However, these studies have some notable limitations and research gaps that give scope for further exploration.\\
Sampath et al.~\cite{63} provided a detailed taxonomy of GAN-based solutions for addressing imbalance in computer vision tasks. It categorizes issues at the image, object, and pixel levels. While the study offered insights into classification, detection, and segmentation challenges, it did not extend its analysis to other crucial domains, such as structured data or non-visual tasks. 
Cole and Khoshgoftaar~\cite{64} focused on the using GANs in tabular datasets, primarily focusing on network traffic classification and financial transactions. Their survey covers the prevalence of CGAN architectures and CNN-based learners having improvements in accuracy. However, their scope was confined to structured data and did not address GAN applications in broader or hybrid frameworks. 
Also, in another study, Cole et al.~\cite{65} adopted a practical analysis of 18 GitHub repositories implementing GAN for imbalanced datasets. Their methodology focused on code structure, library reliance, and best practices for researchers. However, the study lacked a theoretical framework or categorization of GAN techniques across application areas.   
Nayak et al.~\cite{66} examined GAN methodologies in computer vision with a particular focus on image enhancement tasks. Their systematic literature review compared GANs with traditional machine learning and MATLAB-based approaches, proving GANs superior performance in tasks like image denoising and enhancement. While the studyfocused on image quality, it offered limited insights into the broader challenges of dataset imbalance in various domains.
The existing literature reveals critical gaps in the study of GANs in addressing data imbalance. Most reviews have a narrow focus on domains while they lack attention in high-impact areas like healthcare~\cite{67},~\cite{68},~\cite{69}, finance~\cite{70},~\cite{71},~\cite{72},~\cite{73} and cybersecurity~\cite{74},~\cite{75},~\cite{76},~\cite{77} unexplored. Also, the literature fails to provide systematic mapping of GAN techniques, variants, and their application strengths, which are essential for a deeper understanding across domains.

Our methodology addresses these gaps by systematically analyzing studies. This study spans a broader range of application areas like healthcare, finance, cybersecurity, and others. It introduces unique categorization mappings that systematically organize GAN techniques, variants, and their domain-specific applications. It offers a more comprehensive perspective than previous reviews. The research highlights a scope for future research approaches like diffusion models and reinforcement learning by identifying the existing absence of hybridized GAN frameworks. Through these contributions, the study strengthens the existing knowledge and provides a foundation for addressing data imbalance.

\section{Research Approach}

The literature review on GANs addressing data imbalance follows a detailed systematic mapping approach. The primary objective of this methodology is to systematically identify, collect, and categorize literature work. The methodology ensures a transparent and rigorous approach to literature review. Based on a predefined framework developed by Petersen et al.~\cite{78}, the study aims to create a comprehensive overview that allows researchers to understand the current landscape on GANs in handling imbalanced data. The systematic mapping study follows three fundamental procedural stages: (1) formulation of research questions, (2) development of a robust search strategy, and (3) systematic paper filtering process.

\subsection{Research Questions}

Crafting precise and well-defined research investigations is fundamental to conducting a comprehensive literature review in any domain~\cite{79}. Through detailed preliminary research and careful analysis of existing literature, the study developed a strategic set of investigative questions designed to explore and characterize the application of GANs in handling imbalanced datasets. The study strategically segmented into three categories of inquiry, as shown in Table~\ref{tab:research_questions}:

\begin{itemize}
    \item \textbf{Primary Exploration (PE)}: This category establishes the core research objective, exploring the previously used robust strategies in detail and acting as  a clear lens through which the entire investigation is viewed.
    \item \textbf{Focused Investigations (FIs)}: It consists of five targeted research questions, these inquiries comprehensively map out current modeling approaches, uncover underlying patterns, identify critical limitations, and highlight potential research gaps that can inform future research methods.
    \item \textbf{Trend Analysis (TAs)}: It consists of two bibliographic-oriented questions. Firstly, it enables a chronological examination of the research landscape. Secondly, it thoroughly explains the field's evolutionary trajectory and quality assessment.
\end{itemize}

\begin{table}[htbp]
\centering
\caption{Research Investigations}
\label{tab:research_questions}
\begin{tabularx}{\textwidth}{lX}
\hline\hline
\textbf{RI\#} & \textbf{Description} \\ 
\hline\hline
PE1 & How are GANs utilized to address data imbalance challenges in artificial intelligence? \\ 
FI1 & What application domains use GANs for imbalanced data handling? 
\\
FI2 & How are GANs combined with data-level approaches? 
\\ 
FI3 & What algorithmic enhancements are used alongside GANs? 
\\ 
FI4 & Which GAN variants are used for handling the imbalance issue? 
\\ 
FI5 & Which performance metric is used to compare GAN with traditional techniques? \\ 
TA1 & How has research on GANs for imbalanced data evolved over time (publications per year)? \\ 
TA2 & Which journals, conferences, or platforms publish relevant work (publications per venue)? \\ 
\hline
\end{tabularx}
\end{table}

\subsection{Search Strategy}
The search process was designed to follow into three main steps: (1) establishing relevant keywords as the search query, (2) selecting relevant databases, and (3) gathering and refining the results. This systematic approach was designed to include all relevant studies while minimizing irrelevant articles. \\

\textbf{Step 1: Developing the Search Query}
The initial step involved finding keywords and their synonyms from the related field research analysis. These keywords were combined using boolean operators (AND, OR) for an effective search query. Table~\ref{tab:Search Queries} summarizes the details of search strings and the number of results obtained from each database. \\

\begin{table}[h!]
\centering
\caption{Search Queries and Final Results for Each Database}
\label{tab:Search Queries}
\begin{tabular}{|l|p{8cm}|c|}
\hline
\textbf{Database} & \textbf{Search Query} & \textbf{Final Results} \\ 
\hline
\textbf{IEEE Xplore} & 
\textit{(\enquote{Imbalanced Data} OR \enquote{Class Imbalance} OR \enquote{Imbalanced Datasets} OR \enquote{Skewed Data} OR \enquote{Minority Class} OR \enquote{Data Imbalance} OR \enquote{Unbalanced Data} OR \enquote{Uneven Data Distribution}) AND (\enquote{Generative Adversarial Networks} OR \enquote{GAN} OR \enquote{Generative Adversarial Network} OR \enquote{Adversarial Networks} OR \enquote{GANs} OR \enquote{Adversarial Machine Learning} OR \enquote{GAN Models})} & 1,049 \\ 
\hline

\textbf{Scopus} & 
\textit{Article Title, Abstract, Keywords - (\enquote{Imbalanced Data} OR \enquote{Class Imbalance} OR \enquote{Imbalanced Datasets} OR \enquote{Skewed Data} OR \enquote{Minority Class} OR \enquote{Data Imbalance} OR \enquote{Unbalanced Data} OR \enquote{Uneven Data Distribution}) AND (\enquote{Generative Adversarial Networks} OR \enquote{GAN} OR \enquote{Generative Adversarial Network} OR \enquote{Adversarial Networks} OR \enquote{GANs} OR \enquote{Adversarial Machine Learning} OR \enquote{GAN Models})} & 1,166 \\ 
\hline

\textbf{Web of Science} & 
\textit{All Fields - (\enquote{Imbalanced Data} OR \enquote{Class Imbalance} OR \enquote{Imbalanced Datasets} OR \enquote{Skewed Data} OR \enquote{Minority Class} OR \enquote{Data Imbalance} OR \enquote{Unbalanced Data} OR \enquote{Uneven Data Distribution}) AND (\enquote{Generative Adversarial Networks} OR \enquote{GAN} OR \enquote{Generative Adversarial Network} OR \enquote{Adversarial Networks} OR \enquote{GANs} OR \enquote{Adversarial Machine Learning} OR \enquote{GAN Models})} & 679 \\ 
\hline

\textbf{PubMed} & 
\textit{(\enquote{Imbalanced Data} OR \enquote{Class Imbalance} OR \enquote{Imbalanced Datasets} OR \enquote{Skewed Data} OR \enquote{Minority Class} OR \enquote{Data Imbalance} OR \enquote{Unbalanced Data} OR \enquote{Uneven Data Distribution}) AND (\enquote{Generative Adversarial Networks} OR \enquote{GAN} OR \enquote{Generative Adversarial Network} OR \enquote{Adversarial Networks} OR \enquote{GANs} OR \enquote{Adversarial Machine Learning} OR \enquote{GAN Models})} & 147 \\ 
\hline
\end{tabular}
\end{table}

The preliminary exploration of terms also experimented with other phrases, such as substituting \enquote{Generative Adversarial Networks} with \enquote{GANs} or using \enquote{Data imbalance} and \enquote{Skewed Data} in place of \enquote{imbalanced data}. However, these variations often returned irrelevant or overly broad results, such as studies on general classification problems or unrelated domains.\\

\textbf{Step 2: Selecting Databases}
The study used four major academic databases: IEEE Xplore, Scopus, Web of Science, and PubMed. These databases were selected for their broadness in peer-reviewed articles and interdisciplinary content. These platforms are also crucial in including studies on using GANs in imbalanced data applications.

\textbf{Step 3: Refining and Filtering Results}
Some filter criteria were used to select the most relevant studies. These criteria included language like english-only, document types like excluding book chapters, short surveys, and letters, and accessibility like prioritizing open-access publication restrictions. The database-specific exclusions and refinements are summarized as follows: 

\begin{itemize}
\item \textbf{IEEE Xplore:} Excluded six early access articles and two magazines, resulting in a final count of 1,049 articles.
    \item \textbf{Scopus:} Excluded conference reviews (35), reviews (15), book chapters (13), short surveys (2), and letters (1), resulting in 1,301 documents. Language filters removed non-English articles (e.g., 46 Chinese, 4 Korean, and 2 Turkish articles). The exclusion of the book series reduced the count to 1,166 articles.
    \item \textbf{Web of Science:} Excluded 23 early access articles, 17 review articles, three proceeding papers, and one meeting abstract, resulting in 681 documents. Language filters removed two Chinese articles, leaving 679 articles.
    \item \textbf{PubMed:} No specific exclusion criteria were applied, resulting in 147 articles.
\end{itemize}

\textbf{Final Dataset}

After applying these filters and criteria across all databases, a combined total of articles was compiled for review. The Rayyan AI Tool~\cite{80} was used to manage and review the 3,041 retrieved articles. The tool automatically identified and excluded duplicates, reducing the dataset to 1,233 articles. Two authors independently reviewed the abstracts in a blinded manner to minimize bias. Irrelevant studies were excluded based on the abstracts. The authors then merged their independently reviewed results. It yields a final set of 130 relevant articles. Of 130 articles, 30 were excluded based on further analysis, and finally, 100 articles were identified. This two-step process used by authors acted as a dual-authentication mechanism to enhance reliability and ensure consensus.\\

\begin{figure}
    \centering
    \includegraphics[height=0.65\textheight,width=0.85\textwidth]{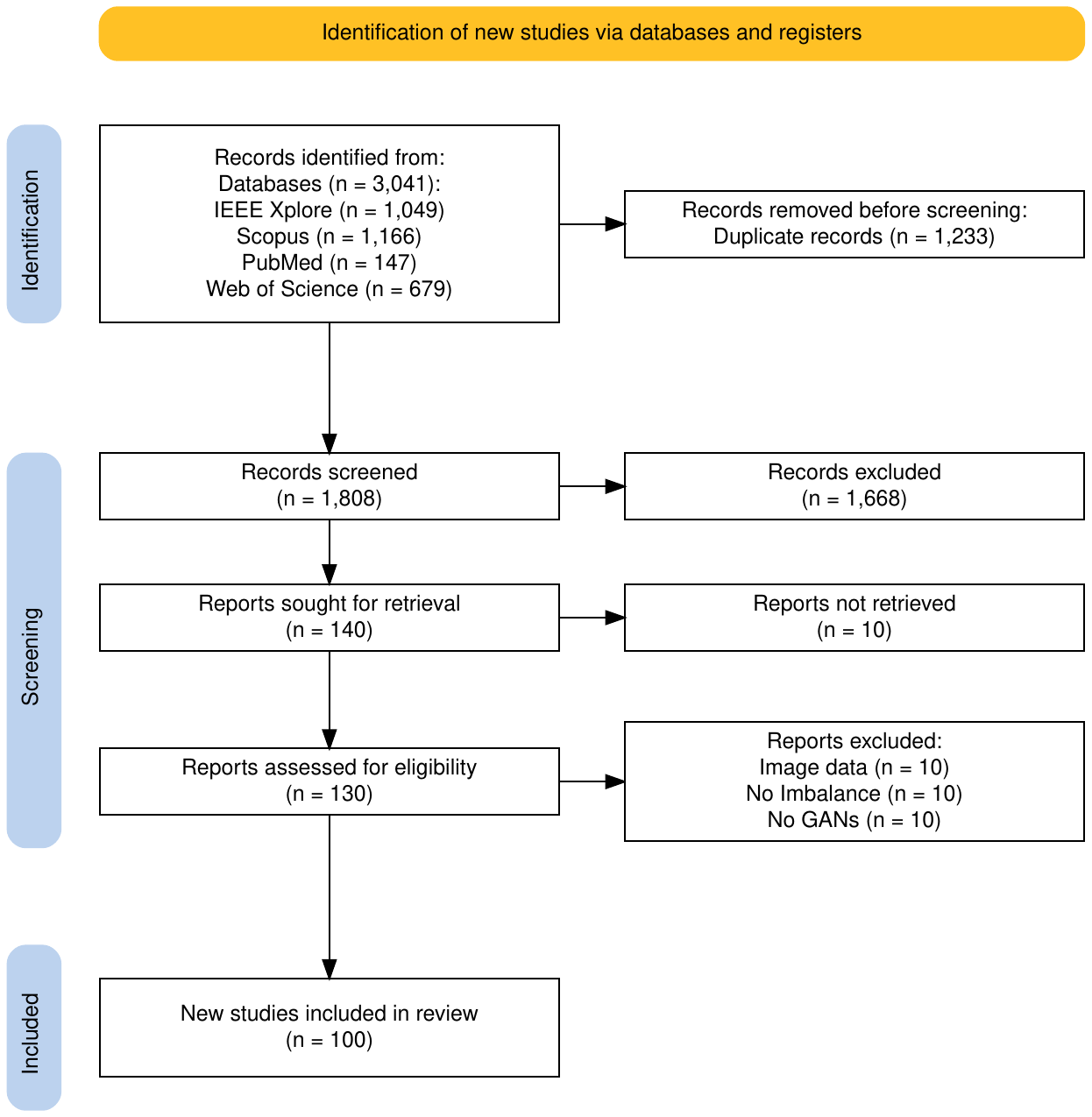}
    \caption{Prisma}
    \label{fig:prisma}
\end{figure}

\textbf{PRISMA Flow Diagram}
Figure~\ref{fig:prisma} visualized the article's identification, screening, eligibility, and inclusion process for articles using the PRISMA (Preferred Reporting Items for Systematic Reviews and Meta-Analyses) flow diagram. The diagram systematically outlines the progression of records through the different phases of the review process, including the number of records identified, duplicates removed, records screened, full articles assessed for eligibility, and studies included in the final review~\cite{81},~\cite{82},~\cite{83}. This approach ensured transparency in the selection process and adherence to systematic review standards.

\subsection{Filtering Process}

A systematic filtering process was used to select relevant and quality studies. The process utilized the following Exclusion Criteria (EC):\\
\begin{itemize}
    \item \textbf{EC1}: The study is published in a venue other than a journal or conference.
    \item \textbf{EC2}: The article is written in a language other than English.
    \item \textbf{EC3}: The article is a literature review rather than original research.
    \item \textbf{EC4}: The full text of the article is not available.
    \item \textbf{EC5}: The article is less than four pages long.
    \item \textbf{EC6}: The study aligns with the search string keywords, but its context does not align with the research objectives.
    \item \textbf{EC7}: The study does excludes tabular data.
    \item \textbf{EC8}: The study does not address imbalanced data as a criterion.
    \item \textbf{EC9}: The study focuses exclusively on Generative Adversarial Networks (GAN) and their variants.
    \item \textbf{EC10}: Others
\end{itemize}

The filtering process began with an initial screening to remove studies based on EC1 and EC2, retaining only journal and conference publications written in English. Following this, title and abstract reviews were conducted to exclude studies meeting EC3 and EC4. These filters selected the contextually relevant works that and focused on original contributions. \\

Further filtering involved a three-stage review process to eliminate papers meeting EC5, EC6, and EC7. These filters focused on all selected studies that were accessible, sufficiently detailed, and directly relevant to tabular data analysis. In the final stage, full-text reviews were performed to remove papers that did not address the core topics of imbalanced data or tabular datasets or that exclusively focused on GANs without broader applicability (EC8 and EC9). EC10 was applied as a catch-all for studies that fell short of quality or relevance benchmarks. This multi-tiered approach ensured that only high-quality studies aligned with the research objectives were included in the final analysis.\\

\section{Results and Discussion}
\subsection{PE1: How are GANs utilized to address data imbalance challenges in artificial intelligence?}
GANs turn up as an effective technique for imbalanced datasets by expanding the training dataset. GANs help machine learning models overcome class distribution issues and improve overall learning performance. \\

~\cite{87} explore the Wasserstein-GAN approach to address data imbalance in fraud detection datasets. The methodology focuses on generating synthetic instances, effectively balancing class distribution, and improving the quality of minority fraudulent transaction data.~\cite{88} propose an imbalanced generative adversarial fusion approach (IFGAN) to address data imbalance challenges in credit scoring applications. The study uses GANs, enabling a semi-supervised learning approach that improves model performance on skewed datasets. The study also introduces a specialized balance module to manage the class distribution problem in financial risk assessment models.~\cite{94} employ conditional tabular GAN (CTGAN) as an oversampling technique to address data imbalance. The research shows GANs potential to improve machine learning model performance on critical health insurance skewed datasets.~\cite{96} employ conditional tabular GAN (CTGAN) to address data imbalance in credit card fraud detection. The study introduces the neighborhood cleaning rule (NCL) algorithm to refine and remove overlapping instances.~\cite{99} explore GANs to solve data imbalance challenges in analytical customer relationship management domains. Techniques like the \enquote{chaoticGAN}, and a hybrid approach combining GAN oversampling with one-class support vector machine undersampling help explore GANs potential to generate high-quality synthetic minority-class samples. The research offers a targeted strategy for improving predictive performance in critical applications like credit card churn, insurance fraud, and loan default prediction.~\cite{101} introduce a new data augmentation model called KCGAN for handling data imbalance in credit card fraud detection. The methodology uses the Kullback‐Leibler divergence as the loss function in the generator for generating the synthetic samples to solve the imbalance issue.~\cite{103} use GANs, SMOTE, and their hybrids to handle a class imbalance in financial fraud datasets. Hybrid approaches like SMOTified-GAN and GANified-SMOTE show comparable and effective performance across varying levels of synthetic fraud data generation.~\cite{108} uses GAN with autoencoders to address class imbalance in credit data. The modified GAN generates new instances from random noise, balancing the dataset by creating synthetic instances for the minority class. Combining autoencoders and GANs provides a novel approach to data augmentation.~\cite{112} utilizes GANs to generate synthetic samples for the minority class in fraud detection datasets. It addresses the issue of data imbalance by providing additional data points for the fraudulent class.~\cite{125} employ GANs to balance the data distribution on the server directly while addressing the issue of data imbalance in federated learning.~\cite{127} highlight the use of unrolled GAN-based oversampling, which captures the data distribution to generate balanced datasets, outperforming traditional oversampling methods in credit card fraud detection.~\cite{129} address the credit card fraud detection problem by applying GANs to handle the imbalanced data issue. They introduce the ccfDetector system, which combines GAN-based data augmentation, deep learning with artificial neural network (ANNs), and feature selection.~\cite{131} explore GANs integrated with SMOTE and VAEs to generate synthetic instances for the minority class in fraud detection imbalanced dataset. It helps to overcome the issue of class imbalance by balancing the dataset and improving classification accuracy.~\cite{133} use GANs to generate synthetic data for minority classes, in bitcoin datasets, such as mining pool entities. It helps handle the class imbalance issue, improving classification outcomes for minority but critical categories.~\cite{134} use GANs to generate synthetic samples of fraudulent transactions. The generative process helps rebalance the dataset before training classifiers.~\cite{137} explicitly uses GANs in combination with SMOTE to generate synthetic samples for the minority class. The methodology helps produce diverse and realistic data while handling the noise and overgeneralization issues associated with traditional oversampling.~\cite{138} used k means CTGAN to generate synthetic instances for the minority class of financial e-commerce datasets while addressing the imbalance problem by producing a dataset with a more uniform distribution preserving the original data's probability distribution.~\cite{139} used GANs to generate synthetic samples to handle class imbalance in credit card fraud detection, a domain where fraudulent activities constitute an extreme minority class.~\cite{143} directly apply GANs to generate synthetic data for the minority class. They aim to improve the availability of minority samples in the credit card fraud detection dataset.~\cite{148} introduce a GAN-based method specifically designed for oversampling the minority class in an imbalanced European credit card dataset.~\cite{149} directly uses K-CGAN to generate synthetic data, aiming to balance imbalanced datasets and improve classifier performance in tasks such as fraud detection. The authors introduce a new loss component based on KL divergence.~\cite{150} use multiple adversarial networks, including vanilla GAN, least squares GAN, Wasserstein GAN, margin adaptive GAN, and Relaxed Wasserstein GAN, to generate pseudo data and address the class imbalance problem for fraud detection.~\cite{154} utilize an improved VAEGAN to generate synthetic minority class samples for oversampling, directly addressing class imbalance challenges in fraud detection. The methodology uses the variational autoencoder along with GAN to address class imbalance.~\cite{155} proposed the Duo-GAN framework, which uses two separate GAN generators, one dedicated to fraudulent instances and the other to legitimate ones. It directly tackles the class imbalance problem in fraud detection.~\cite{156} employs GANs to generate synthetic samples for the minority class, balancing the imbalanced bankruptcy datasets.~\cite{157} use GANs to generate synthetic minority class samples in credit card fraud detection datasets. Also, they argued that GANs are more stable and effective than traditional oversampling algorithms.~\cite{161} use GANs to oversample the minority class (defaulted bonds) to address the class imbalance problem. They proposed a GAN and CNN approach to handle the class imbalance.~\cite{166} employ GANs as an oversampling technique to generate synthetic fraud samples. They use contrastive loss or triplet loss as auxiliary loss functions, and the effect of discriminator rejection sampling on synthetic data generation.~\cite{167} proposed cycle-consistent adversarial networks (CycleGAN) that generate synthetic data for the minority class, addressing data imbalance in fraud detection. They argued that Cycle-GAN improves the performance of the fraud detection models.~\cite{170} use GANs as an oversampling method for the commercial bank customer dataset. This methodology shows better application value through experiments and comparison with traditional oversampling methods.~\cite{173} propose the auxiliary conditional tabular generative adversarial network (ACTGAN), tailored for tabular data, and prove effective in generating realistic synthetic samples for the minority class. They also designed ResNet-LSTM for the feature selection process.~\cite{176} provide a unique hybrid approach that combines GAN-generated data with undersampling to improve minority-majority class separability in finance data.~\cite{177} uses CGAN to oversample minority classes in the fraud detection dataset.~\cite{178} use GANs to oversample fraudulent transactions, with an innovative focus on class boundary diversity.~\cite{181} constructed a GAN-stacking model for classification prediction. GANs are used for data augmentation to address the imbalance in fraud datasets. \\

~\cite{84},~\cite{183} use CTGAN to generate synthetic data to augment an imbalanced dataset of medication use in critically ill patients. It helps balance the minority class to improve predictive modeling performance. The methodlogies explored the fluids overload prediction and intensive care unit (ICU) settings dataset and improved predictive models for irregular temporal data.~\cite{85} propose an EC-WGAN approach that uses GANs to augment minority classes in ECG datasets synthetically. The hybrid method simultaneously trains the GAN and classifier. It also optimizes performance and reduces training overhead for detecting shockable cardiac rhythms.~\cite{86} explore GAN-based oversampling techniques as solutions for addressing data imbalance in Parkinson's disease dementia patient datasets. By comparing advanced generative approaches with traditional resampling methods like SMOTE and random over sampling. The findings highlight GANs usefulness in balancing the minority class.~\cite{89} introduce actGAN to address the imbalance in vital statistics datasets with rare outcomes like early stillbirth. The method aims for an acceptable predictive model performance when dealing with extremely skewed and limited medical data. The study shows GANs potential for improving machine learning models in challenging, low-prevalence medical scenarios.~\cite{97} introduce the gradually generative adversarial network (GradGAN) to solve data imbalance challenges in machine learning. It progressively generates synthetic minority class samples and incrementally balances class distribution. The method also offers a dynamic approach to addressing healthcare-skewed datasets.~\cite{102} uses GAN to generate the synthetic data for minority class in addressing the class imbalance in a cancer intracellular signaling dataset. The research focuses on GAN's role in oversampling and improving classification accuracy.~\cite{104} discusses conditional tabular generative adversarial networks (CTGANs) for generating synthetic categorical clinical data to address class imbalance. The methodology aimed at improving predictive performance in ML models for cardiovascular disease (CVD) risk prediction.~\cite{106} proposed G-GAN, a GAN-based oversampling method designed for diabetes imbalanced data. The methodology uses prior knowledge of the minority class by estimating a Gaussian distribution of minority samples to guide the latent space of GAN.It utilizes the idea of bagging to avoid overfitting while generating dispersive minority samples.~\cite{107} use EvaGoNet as part of the feature engineering framework within the Gaussian Mixture Variational Autoencoder (GMAVE) decoder module. WGAN with Gradient Penalty handles decoding and enables better latent feature representation for imbalanced bioOMIC structured datasets.~\cite{111} use GAN to generate synthetic sequence data that resembles real data. The synthetic data enables improved ML model performance for classifying biological sequences.~\cite{126} present a GAN-based approach to asthma diagnosis using feature selection, data augmentation, and extreme gradient boosting (XgBoost) for classification.~\cite{145} use tabular GANs to generate synthetic samples for the minority class in imbalanced survival datasets.~\cite{159} proposed HAR-CTGAN to generate synthetic data to handle class imbalance in human activity recognition data.It focuses on the synthesizing continuous features, such as real-number data recorded from various sensors.~\cite{160} used GANs to augment and synthesize data for balancing the cardiovascular disease prediction dataset. They use GAN and SVM-based feature selection to handle the class imbalance problem.~\cite{162} use GANs to generate synthetic electronic health record (EHR) data. The synthetic data generated by GANs provide balanced datasets and improve the training of ML algorithms by providing additional instances for minority classes (e.g., specific length of stay LOS categories).~\cite{179} use GANs for generating synthetic minority instances for predicting diabetes datasets. However, they argue that GANs are less effective compared to simpler undersampling techniques.~\cite{182} introduce BM-WGAN for the synthetic data generation to balance datasets with biological data. The methodology aims to use the bootstrap method for the input to the generator. \\

~\cite{95} use a Wasserstein conditional GAN with a penalty (WCGAN-GP) to address data imbalance by generating synthetic network intrusion detection system (NIDS) tabular data. The study uses a methodology that generates high-quality minority-class samples. It focuses on improving detection rates and minimizing false positives, with the transformative potential of generative techniques in handling imbalanced NIDS data challenges.~\cite{98} intoduce HT-Fed-GAN, a groundbreaking federated generative model that uses GANs to address data imbalance challenges in complex privacy preserving tabular datasets. Some advanced techniques like variational Bayesian Gaussian mixture modeling and conditional sampling are used to generate synthetic data that balances multimodal and categorical distributions.~\cite{100} introduce a modified conditional GAN (MCGAN) to address class imbalance in intrusion detection systems. The method balances dataset distributions and improves predictive model performance.~\cite{109} employ the WGAN-GP model to generate the synthetic samples for sparse attack classes in the NSL-KDD dataset. The WGAN-GP improves the recognition accuracy of rare network attack packets, which are otherwise challenging to classify due to their sparsity.~\cite{113} utilizes a conditional tabular GAN (CTGAN) using the UNSW-NB 15 dataset, which has fewer instances of intrusion compared to normal network behavior. CTGAN balances the dataset and improves the classification performance of intrusion detection models.~\cite{130} propose CT-GAN (Conditional Table GAN) architecture to address the challenge of limited data availability in malware detection. It uses continuous and categorical variables in a unified framework, unlike existing GAN approaches that treat them separately.~\cite{136} highlights CTGAN, a conditional GAN architecture for handling the tabular data and generating the minority class instances. Using CTGAN, they tried to combat skewed class distributions in intrusion detection.~\cite{140} use CTGAN architecture of GAN combined with a variant of variational autoencoder (TVAE) as a synthetic data generator, for handling imbalanced datasets. They explore the class imbalance problem using GANs in the malware detection and wafer manufacturing domain.~\cite{142} study various conditional GAN architectures, and address the data imbalance problem in malware detection dataset and improve classifier performance.~\cite{151} use WGAN-DIV as part of a hybrid oversampling model to generate synthetic data for the minority class (malicious packets) in NIDS. It addresses the class imbalance issue by increasing the representation of attack instances in the training dataset.~\cite{153} focus on synthetic tabular data generation; the paper mentions a comparison of TABGAN and CTGAN. However, the approach addresses class imbalance during the data generation process in intrusion detection systems.~\cite{169} proposed a convergent WGAN to tackle class imbalance by balancing the network threat detection dataset. They argue that CWGAN not only improves the performance of the minority class through oversampling but also improves the training stability of WGAN. \\

~\cite{172} propose SC-GAN, which uses conditional GAN to model sequential data, generating synthetic minority-class samples for student performance prediction.~\cite{115} use GANs as one of the data augmentation techniques for improving classification performance on imbalanced weather data. Hence, GANs are used to balance the dataset and improve the model's ability to classify rare but critical weather events.~\cite{116} apply a GAN-based approach to address data imbalance using SA-CTGAN, which generates synthetic fault samples in electrical substations to balance the dataset. The self-attention mechanism helps maintain the correlation between the data features, ensuring that the generated samples are realistic and relevant to the task.~\cite{117} apply CTGAN to generate a synthetic overbreak dataset to solve the issue of data shortage and imbalance issue. The model generates synthetic data that retains the properties of the original dataset, improving its diversity and handling data imbalance for overbreak prediction in tunnel blasting.~\cite{144} study uses a hybrid GAN architecture PacGAN to generate synthetic samples for minority classes, balancing datasets for recommendation systems. They integrate conditional Wasserstein GAN with a gradient penalty and additional mechanisms like auxiliary classifier loss for realistic and relevant synthetic data.~\cite{163} uses a CWGAN architecture to generate quality backorder samples to address the issue of imbalanced data in product backorder prediction.\\

~\cite{90} use CWGAN as an oversampling technique to address data imbalance challenges. Using GAN, they generate synthetic instances for minority instances and balance skewed datasets across multiple real-world domain datasets.~\cite{91} explore the class imbalance by introducing a novel borderline conditional generative adversarial network (BCGAN) approach. By targeting minority class data near decision boundaries, the model uses GAN to generate synthetic samples that enhance model performance on imbalanced datasets.~\cite{92} introduce the conditional generative adversarial imputation network (CGAIN) to address data imbalance through missing data imputation. The study provides a targeted solution for handling diverse and skewed datasets by conditioning the GAN generation process on class labels.~\cite{93} use the advanced conditional Wasserstein generative adversarial network (ACWGAN) to solve multi-class data imbalance challenges. The study uses GAN to generate synthetic samples across multiple minority classes and offers an effective approach to balancing complex, skewed datasets.~\cite{105} describe a GAN-based model (RVGAN-TL) that uses a variational autoencoder (VAE) to improve GAN-generated data quality. It introduces techniques such as similarity measure loss and a roulette wheel selection method to rebalance the dataset.~\cite{110} propose a hybrid sampling model for various domain tabular data to oversample minority class samples. This approach uses a GAN variant (TWGAN-GP) that combines clustering with the GAN to help balance the dataset and improve classification performance.~\cite{114} propose CTGAN-based auxiliary classifier GAN (ACCTGAN), which generates synthetic data for the sparser minority class. Before generating the sparser minority class, preprocessing or undersampling carried out by using clustering. The methodology improves the model's ability to classify the minority class of various domain datasets correctly.~\cite{118} propose I-GAN, which improves traditional GANs by using additional density distribution information for the minority class in various domain datasets. The methodology allows the generator to create samples aligned with both local and global distributions. This approach resolves a common limitation of oversampling methods that only rely on local density.~\cite{119} introduces a discriminative boundary generation framework (BoG) based on adversarial training to generate synthetic boundary outliers and augment the dataset to provide critical information. The methodology handles the class imbalance by expanding the dataset with boundary outliers and boundary-normal data.~\cite{120} introduce Generative Adversarial Minority Enlargement (GAME) to generate synthetic samples for the minority class in various domain datasets. GAME  extends the sampling margin during the data generation process by adjusting the parameters of a local linear model. It allows the synthetic samples better to represent the unsampled region of the minority class.~\cite{121} proposed the ConvGeN model that combines convex space learning with deep generative techniques to generate synthetic samples for the minority class of various domain datasets. The methodology focuses on learning coefficients for convex combinations of the minority samples and helps generate synthetic samples distinct from the majority class.~\cite{122} employ an adaptive weighting GAN (AWGAN) approach to generate synthetic samples for the minority class by adapting to the real data distribution of various domain datasets. The key innovation in methodology lies in adaptive weighting and density-based analysis to resolve issues like class overlap, intra-class imbalance, and noise. It allows AWGAN to generate synthetic instances that better represent the minority class and improve overall data balance.~\cite{123} propose an improved GAN and biased loss combined model (VGAN-BL) to generate high-quality synthetic samples for the minority class in various domain imbalanced datasets. The methodology uses variational autoencoders to generate latent vectors from a posterior distribution, which are then used as inputs to the GAN. Additionally, the generator is enhanced by introducing Kullback-Leibler similarity measurement loss, improving the quality of minority samples generated.~\cite{124} introduce two novel GAN-based oversampling techniques, GAN-based Oversampling (GBO) and Support Vector Machine-SMOTE-GAN (SSG), to handle class imbalance for various domain datasets. These methods generate realistic minority class samples using GAN while addressing overlapping and distribution fidelity that are common in traditional oversampling methods.~\cite{128} introduce MIX-NOISE GAN, which addresses the issue of mode collapse in GANs when handling multi-cluster distributions of positive samples in binary classification with various domain imbalanced datasets. It uses a Gaussian Mixture Model to generate noise that aligns with the distribution of positive samples, combined with traditional random noise as the input to the generator.~\cite{132} used GANs in conjunction with SMOTE to transform synthetic samples generated by SMOTE into more realistic minority class data. They addresses the class imbalance in various domain datasets by improving the quality of synthetic samples for the minority class.~\cite{135} propose SB-GAN to address the class imbalance for various domain datasets by considering noisy and borderline samples. They aim at learning better class distributions that are not distorted by the existence of outliers.~\cite{141} uses GANs to generate new data for the minority labels in multilabel datasets in various domains. They handle the imbalance problem by resampling the minority classes to improve model performance.~\cite{146} utilize CTGAN architecture to generate synthetic samples for the minority class in various domain datasets. They incorporate Gaussian Mixture Models as a clustering step before GAN-based oversampling, which improves the GAN's ability to focus on specific class distributions.~\cite{147} introduce a novel adversarial training-based approach for imbalanced datasets, extending GAN principles. Instead of focusing only on oversampling minority classes, it uses a cost-sensitive classifier and semi-supervised learning, addressing data imbalance in a broader sense.~\cite{152} proposed the LPGOS model that incorporates a GAN-based oversampling approach that uses a variational encoder to capture the posterior distribution of latent variables.~\cite{158} propose the use of CTGAN-MOS to generate synthetic samples for the minority class in various domain datasets. They use a coin-throwing algorithm that removes noise from the augmented data.~\cite{164} use CGAN to approximate the minority class distribution in various domain datasets. They highlight CGAN's strength in capturing global patterns rather than relying only on local relationships.~\cite{165} propose a GAN-based solution to augment the minority class while introducing an outlier detection mechanism via the discriminator to improve the quality of synthetic data in various domain datasets.~\cite{168} propose CWGAN-GP, which adds auxiliary conditional information to WGAN-GP to address the imbalance issue by generating realistic synthetic samples for various domain datasets.~\cite{171} propose CTAB-GAN to address imbalance through a custom-designed conditional vector that encodes skewed distributions, generating realistic minority-class samples in various domain datasets.~\cite{174} propose BIDC1 and BIDC2 two binary imbalanced data classification models. The models use GANs to address the diversity and distributional representation of synthetic minority samples in various domain datasets.~\cite{175} use a dual-discriminator architecture and focus on borderline regions, highlighting a unique GAN application for various domain imbalanced datasets.~\cite{180} proposed a method called OAGAN to generate synthetic minority class samples that mimic the actual data distribution. In this study, CTGAN is combined with adaptive synthetic sampling to consider different weights for minority class samples, improving the diversity of the generated data. \\

\subsection{FI1:What application domains use GANs for imbalanced data handling?}

GANs have found applications across several domains to address the challenges of imbalanced data, as shown in Figure~\ref{fig:pdf_figure}. These 100 reviewed articles have domains that include healthcare, finance, cybersecurity, mixed domain datasets and other unique domains. \\
\begin{sidewaysfigure}
    \centering
    \includegraphics[width=1.0\textwidth]{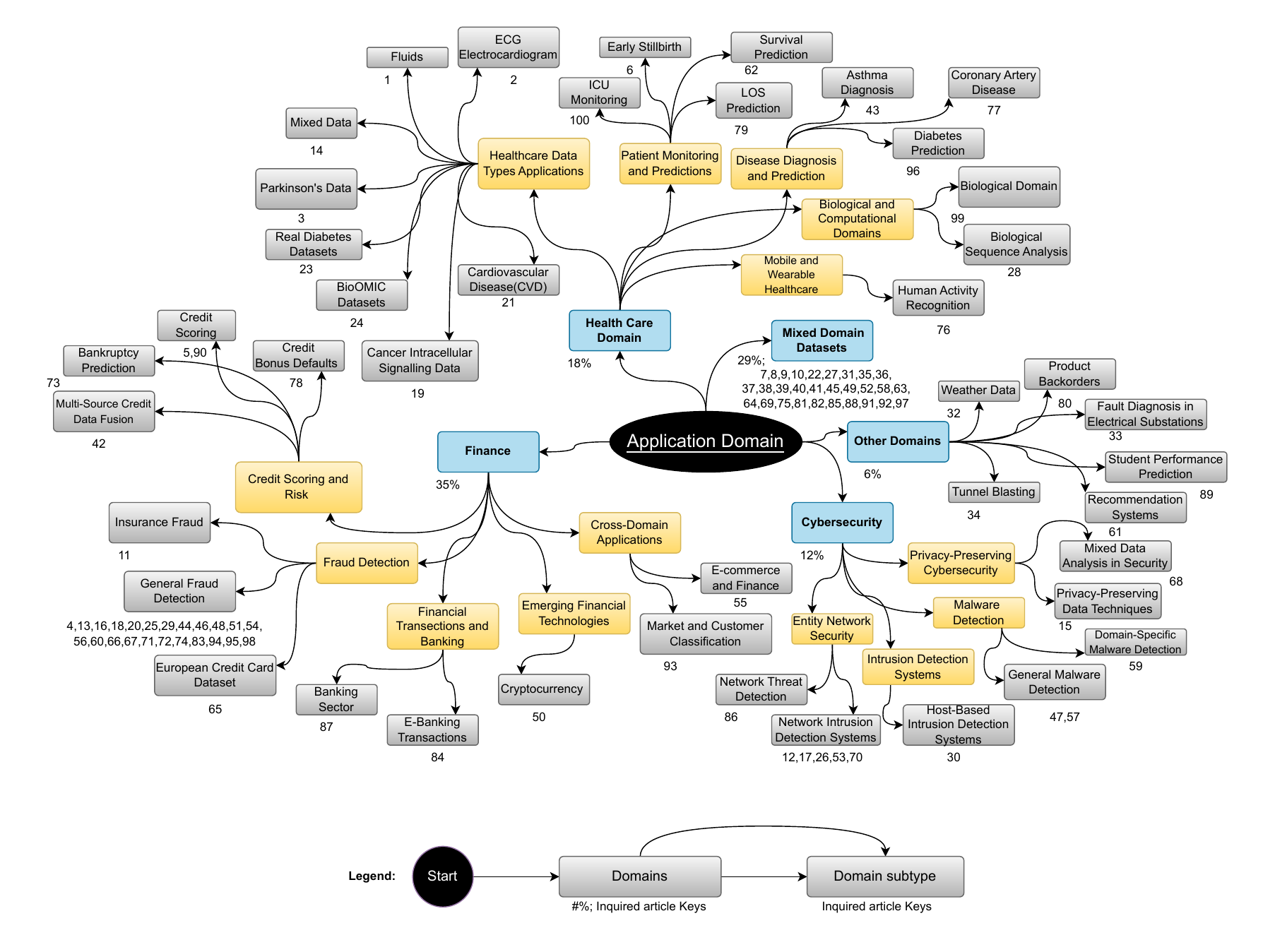}
    \caption{Categorization of domains introduced or analyzed in the examined articles by their corresponding keys}
    \label{fig:pdf_figure}
\end{sidewaysfigure}
Starting with finance, which constitutes the largest share (35\%), GANs address critical challenges such as fraud detection, credit scoring, bankruptcy prediction, emerging financial technologies, and others. These include analyzing European credit card transaction data, detecting e-banking fraud, modeling cryptocurrency markets, and classifying market and costumer-domain applications. GANs excel in processing sparse and sensitive financial data, offering critical risk management and decision-making benefits. \\
Next are mixed domain datasets, which account for 29\% of the applications. GANs in this category are applied to problems involving the integration of diverse data types. These datasets span a variety of multiple domains. \\
Healthcare, representing 18\% of the reviewed works, uses GANs to improve patient care and medical research. Key applications include predicting diseases such as diabetes, coronary artery disease, and asthma, as well as survival prediction, early stillbirth detection, ICU monitoring, and length of stay prediction. Also, some unique applications domains include biological and computation, mobile and wearable healthcare. GANs enable synthetic data generation for high-dimensional imbalanced medical datasets like ECG recordings and Parkinson's data, supporting improved diagnostics and predictions. \\
Cybersecurity accounts for 12\% of the applications where GANs are used to strengthen data security and privacy. GANs play a key role in intrusion detection, malware detection, entity network security, and privacy-preserving cybersecurity techniques. Applications include host-based intrusion detection, network threat detection, and domain-based malware prevention, indicating GANs are used to handle evolving security threats in imbalanced cybersecurity datasets. \\
Lastly, other interesting domains make up 6\% of the reviewed applications, showcasing GANs versatility in less traditional imbalanced areas. These include student performance prediction, tunnel blasting recommendation, fault diagnosis in electrical substations, product backorder prediction, weather data modeling, and recommendation systems. These diverse use cases reflect GANs ability to address unique challenges across various fields. \\
The progression across these domains shows GANs applicability and transformative potential in addressing data imbalance. GANs offer inventive solutions tailored to the specific needs of each domain. The versatility underlines their growing importance in modern machine learning applications. \\

\subsection{FI2:How are GANs combined with data-level approaches?}

Using GANs with data-level approaches has gained attention in addressing class imbalance problems. Two main categories, oversampling and hybrid sampling, have appeared as the primary strategies by GANs, as detailed in Tables~\ref{tab:article_keys1}~\&~\ref{tab:article_keys2}. \\

\begin{table}[ht]
\centering
\resizebox{\textwidth}{!}{
\begin{tabular}{|c|c|c|p{8cm}|c|c|c|}
\hline
\textbf{Article Key (ID)} & \textbf{Ref} & \textbf{Sampling} & \textbf{Approach/Methodology} \textbf{Category} & \textbf{Year} & \textbf{Venue} \\
\hline
1 & \cite{84} &  Hybrid Sampling & SMOTE and GAN & 2024 & Journal \\
2 & \cite{85} & Oversampling & Gan Based Oversampling & 2022 & Journal \\
3 & \cite{86} & Oversampling & Gan Based Oversampling & 2023 & Conference\\
4 & \cite{87} & Oversampling & Gan Based Oversampling & 2020 & Journal \\
5 & \cite{88} & Oversampling & Gan Based Oversampling & 2019 & Journal \\
6 & \cite{89} & Oversampling & Gan Based Oversampling & 2020 & Journal\\
7 & \cite{90} & Oversampling & Gan Based Oversampling & 2021 & Journal\\
8 & \cite{91} & Oversampling & Gan Based Oversampling & 2021 & Journal\\
9 & \cite{92} & Oversampling & Gan Based Oversampling & 2021 & Journal\\
10 & \cite{93} & Oversampling & Gan Based Oversampling & 2022 & Journal\\
11 & \cite{94} & Oversampling & Gan Based Oversampling & 2022 & Journal\\
12 & \cite{95} & Oversampling & Gan Based Oversampling & 2022 & Journal\\
13 & \cite{96} & Hybrid Sampling & NCL(Neighbour hood cleaning) & 2023 & Journal\\
14 & \cite{97} & Oversampling & Gan Based Oversampling & 2023 & Journal\\
15 & \cite{98} & Oversampling & Gan Based Oversampling & 2022 & Journal\\
16 & \cite{99} & Hybrid Sampling & Gan and class support vector machine undersampling & 2022 & Journal\\
17 & \cite{100} & Oversampling & Gan Based Oversampling & 2023 & Journal\\
18 & \cite{101} & Oversampling & Gan Based Oversampling & 2023 & Journal\\
19 & \cite{102} & Oversampling & Gan Based Oversampling & 2023 & Journal\\
20 & \cite{103} & Hybrid Sampling & SMOTE and GAN & 2023 & Journal\\
21 & \cite{104} & Oversampling & Gan Based Oversampling & 2023 & Journal\\
22 & \cite{105} & Oversampling & Gan Based Oversampling & 2023 & Journal\\
23 & \cite{106} & Oversampling & Gan Based Oversampling & 2023 & Journal\\
24 & \cite{107} & Oversampling & hybrid feature-level & 2023 & Journal\\
25 & \cite{108} & Oversampling & Gan Based Oversampling & 2023 & Journal\\
26 & \cite{109} & Oversampling & Gan Based Oversampling & 2023 & Journal\\
27 & \cite{110} & Hybrid Sampling & Gan oversampling and undersampling & 2023 & Journal\\
28 & \cite{111} & Oversampling & Gan Based Oversampling & 2023 & Journal\\
29 & \cite{112} & Oversampling & Gan Based Oversampling & 2023 & Journal\\
30 & \cite{113} & Oversampling & Gan Based Oversampling & 2023 & Journal\\
31 & \cite{114} & Hybrid Sampling & Gan Based Oversampling and clustering(undersampling) & 2024 & Journal\\
32 & \cite{115} & Oversampling & Gan Based Oversampling & 2024 & Journal\\
33 & \cite{116} & Oversampling & Gan Based Oversampling & 2024 & Journal\\
34 & \cite{117} & Oversampling & Gan Based Oversampling & 2023 & Journal\\
35 & \cite{118} & Oversampling & Gan and Density Based Oversampling & 2024 & Journal\\
36 & \cite{119} & Hybrid Sampling & SMOTE and GAN with Boundary Outlier Detection & 2023 & Journal\\
37 & \cite{120} & Oversampling & Gan Based Oversampling & 2024 & Journal\\
38 & \cite{121} & Oversampling & Gan Based Oversampling & 2024 & Journal\\
39 & \cite{122} & Oversampling & Gan Based Oversampling & 2024 & Journal\\
40 & \cite{123} & Oversampling & Gan Based Oversampling & 2023 & Journal\\
41 & \cite{124} & Hybrid Sampling & SVM, SMOTE and GAN & 2024 & Journal\\
42 & \cite{125} & Oversampling & Gan Based Oversampling & 2023 & Journal\\
43 & \cite{126} & Oversampling & Gan Based Oversampling & 2024 & Journal\\
44 & \cite{127} & Oversampling & Gan Based Oversampling & 2022 & Conference\\
45 & \cite{128} & Oversampling & Gan Based Oversampling & 2023 & Conference\\
46 & \cite{129} & Oversampling & Gan Based Oversampling & 2023 & Conference\\
47 & \cite{130} & Oversampling & Gan Based Oversampling & 2023 & Conference\\
48 & \cite{131} & Oversampling & Gan Based Oversampling & 2024 & Conference\\
49 & \cite{132} & Hybrid Sampling & SMOTE and GAN & 2022 & Journal\\
50 & \cite{133} & Oversampling & Gan Based Oversampling & 2020 & Conference\\
\hline
\end{tabular}
}
\caption{Mapping of Article Keys to Categories, Methodologies, and Reference Details}
\label{tab:article_keys1}
\end{table}

\begin{table}[ht]
\centering
\resizebox{\textwidth}{!}{
\begin{tabular}{|c|c|c|p{8cm}|c|c|c|}
\hline
\textbf{Article Key (ID)} & \textbf{Ref} & \textbf{Sampling} & \textbf{Approach/Methodology} \textbf{Category} & \textbf{Year} & \textbf{Venue} \\
\hline
51 & \cite{134} & Oversampling & Gan Based Oversampling & 2023 & Journal\\
52 & \cite{135} & Oversampling & Gan Based Oversampling & 2023 & Conference\\
53 & \cite{136} & Oversampling & Gan Based Oversampling & 2022 & Conference\\
54 & \cite{137} & Hybrid Sampling & SMOTE and GAN & 2023 & Journal\\
55 & \cite{138} & Oversampling & Gan Based Oversampling & 2021 & Conference\\
56 & \cite{139} & Oversampling & Gan Based Oversampling & 2022 & Conference\\
57 & \cite{140} & Oversampling & Gan Based Oversampling & 2023 & Conference\\
58 & \cite{141} & Oversampling & Gan Based Oversampling & 2023 & Conference\\
59 & \cite{142} & Oversampling & Gan Based Oversampling & 2022 & Conference\\
60 & \cite{143} & Oversampling & Gan Based Oversampling & 2021 & Conference\\
61 & \cite{144} & Hybrid Sampling & GAN and Auxillary Classifier & 2021 & Journal\\
62 & \cite{145} & Oversampling & Gan Based Oversampling & 2023 & Conference\\
63 & \cite{146} & Oversampling & Gan Based Oversampling & 2023 & Conference\\
64 & \cite{147} & Oversampling & Gan Based Oversampling & 2023 & Conference\\
65 & \cite{148} & Oversampling & Gan Based Oversampling & 2022 & Conference\\
66 & \cite{149} & Oversampling & Gan Based Oversampling & 2022 & Conference\\
67 & \cite{150} & Oversampling & Gan Based Oversampling & 2018 & Conference\\
68 & \cite{151} & Oversampling & Gan Based Oversampling & 2020 & Conference\\
69 & \cite{152} & Oversampling & Gan Based Oversampling & 2020 & Conference\\
70 & \cite{153} & Oversampling & Gan Based Oversampling & 2023 & Conference\\
71 & \cite{154} & Oversampling & Gan Based Oversampling & 2023 & Journal\\
72 & \cite{155} & Oversampling & Gan Based Oversampling & 2021 & Journal\\
73 & \cite{156} & Oversampling & Gan Based Oversampling & 2023 & Conference\\
74 & \cite{157} & Oversampling & Gan Based Oversampling & 2021 & Conference\\
75 & \cite{158} & Oversampling & Gan Based Oversampling & 2023 & Journal\\
76 & \cite{159} & Oversampling & Gan Based Oversampling & 2022 & Conference\\
77 & \cite{160} & Oversampling & Gan Based Oversampling & 2023 & Conference\\
78 & \cite{161} & Oversampling & Gan Based Oversampling & 2024 & Journal\\
79 & \cite{162} & Oversampling & Gan Based Oversampling & 2023 & Conference\\
80 & \cite{163} & Hybrid Sampling & Gan+Randomized Undersampling (RUS) & 2022 & Conference\\
81 & \cite{164} & Oversampling & Gan Based Oversampling & 2017 & Journal\\
82 & \cite{165} & Oversampling & Gan Based Oversampling & 2019 & Journal\\
83 & \cite{166} & Hybrid Sampling & Discriminator Rejection Sampling (DRS) with GAN-based oversampling & 2020 & Conference\\
84 & \cite{167} & Oversampling & Gan Based Oversampling & 2020 & Conference\\
85 & \cite{168} & Oversampling & Gan Based Oversampling & 2019 & Journal\\
86 & \cite{169} & Oversampling & Gan Based Oversampling & 2021 & Journal\\
87 & \cite{170} & Oversampling & Gan Based Oversampling & 2020 & Conference\\
88 & \cite{171} & Oversampling & Gan Based Oversampling & 2021 & Conference\\
89 & \cite{172} & Oversampling & Gan Based Oversampling & 2021 & Journal\\
90 & \cite{173} & Oversampling & Gan Based Oversampling & 2022 & Journal\\
91 & \cite{174} & Oversampling & Gan Based Oversampling & 2022 & Journal\\
92 & \cite{175} & Hybrid Sampling & Gan and Borderline focused Based Oversampling & 2022 & Journal\\
93 & \cite{176} & Hybrid Sampling & GAN and Undersampling (adaptive, neighborhood-based) & 2022 & Journal\\
94 & \cite{177} & Oversampling & Gan Based Oversampling & 2022 & Conference\\
95 & \cite{178} & Oversampling & Gan Based Oversampling & 2022 & Conference\\
96 & \cite{179} & Oversampling & Gan Based Oversampling & 2022 & Conference\\
97 & \cite{180} & Hybrid Sampling & GAN-based oversampling+Adaptive Synthetic Sampling & 2023 & Conference\\
98 & \cite{181} & Oversampling & Gan Based Oversampling & 2023 & Conference\\
99 & \cite{182} & Oversampling & Gan Based Oversampling & 2024 & Journal\\
100 & \cite{183} & Hybrid Sampling & SMOTE and GAN & 2023 & Journal\\
\hline
\end{tabular}
}
\caption{Mapping of Article Keys to Categories, Methodologies, and Reference Details}
\label{tab:article_keys2}
\end{table}

GAN-based oversampling is the most used approach, as many studies use GANs as the primary technique to generate synthetic samples for the minority class. These methods aim to improve data balance by synthesizing data points that capture the distribution of the minority class. Advanced variations, such as density-based oversampling, clustering-based oversampling and borderline-focused techniques, have further improved the quality of the generated samples, which help deal with challenging datasets. In some cases, oversampling is used with adaptive undersampling techniques to address majority-class redundancy and improve classification performance. \\

On the other hand, hybrid sampling approaches combine GANs with traditional techniques like the synthetic minority oversampling technique (SMOTE) or undersampling to synthesize more robust data points. For instance, methods like SMOTE combined with GANs use the diversity of GAN-generated samples while reducing duplication of data associated with SMOTE. Other studies have explored the combination of GANs with randomized undersampling or adaptive sampling strategies to address overlapping distributions and noisy data. Some advanced hybrid methods, such as Discriminator Rejection Sampling (DRS) combined with GANs, focus on refining the synthetic data generation. \\
While oversampling dominates the research landscape, with 83 studies focusing on GAN-based oversampling, hybrid approaches (17 studies) are gaining momentum due to their ability to handle complex challenges like class overlap and data noise. This trend is visible in recent years (2022-2024) as hybrid techniques have shown their potential to deliver balanced datasets and improved classifier performance. Hence, combining GANs with data-level approaches offers a powerful tool set for addressing imbalanced datasets. 

\subsection{FI3:What algorithmic enhancements are used alongside GANs?}
GANs are often augmented with a range of algorithmic enhancements when handling problems related to imbalanced data. The reviewed articles span neural network architecture, generative models, classification-ensemble methods, data handling methods, loss functions and regularization, feature engineering, learning paradigms, optimization techniques, specialized detection-security, and theoretical approaches, as shown in Figure~\ref{fig:pdf_figure}. These approaches collectively improve the performance and applicability of GAN across various domains. \\

A key foundation for these algorithms is highlighted by neural network architecture, which allows GANs to model complex data patterns. Variants such as Convolutional Neural Networks (CNNs), Recurrent Neural Networks (RNNs), and hybrid models that combine feed-forward networks with CNNs allow GANs to generate high-dimensional data. Autoencoder variants, including Variational Autoencoders (VAEs) and Extreme Learning Machine Autoencoders, further improve GANs by refining feature extraction. These techniques help in the synthesis of diverse and high-quality samples. \\
Building on this foundation, generative models extend the capabilities of GANs to handle specific challenges. Frameworks, such as PACGAN, address mode collapse by ensuring diverse sample generation, while dual-discrimination GANs improve class separation and sample quality. Auxiliary classifiers, integrated into GAN architectures, align the synthetic data generation process with classification objectives, improving the handling of minority classes. \\
\begin{sidewaysfigure}
    \centering
    \includegraphics[width=1.1\textwidth]{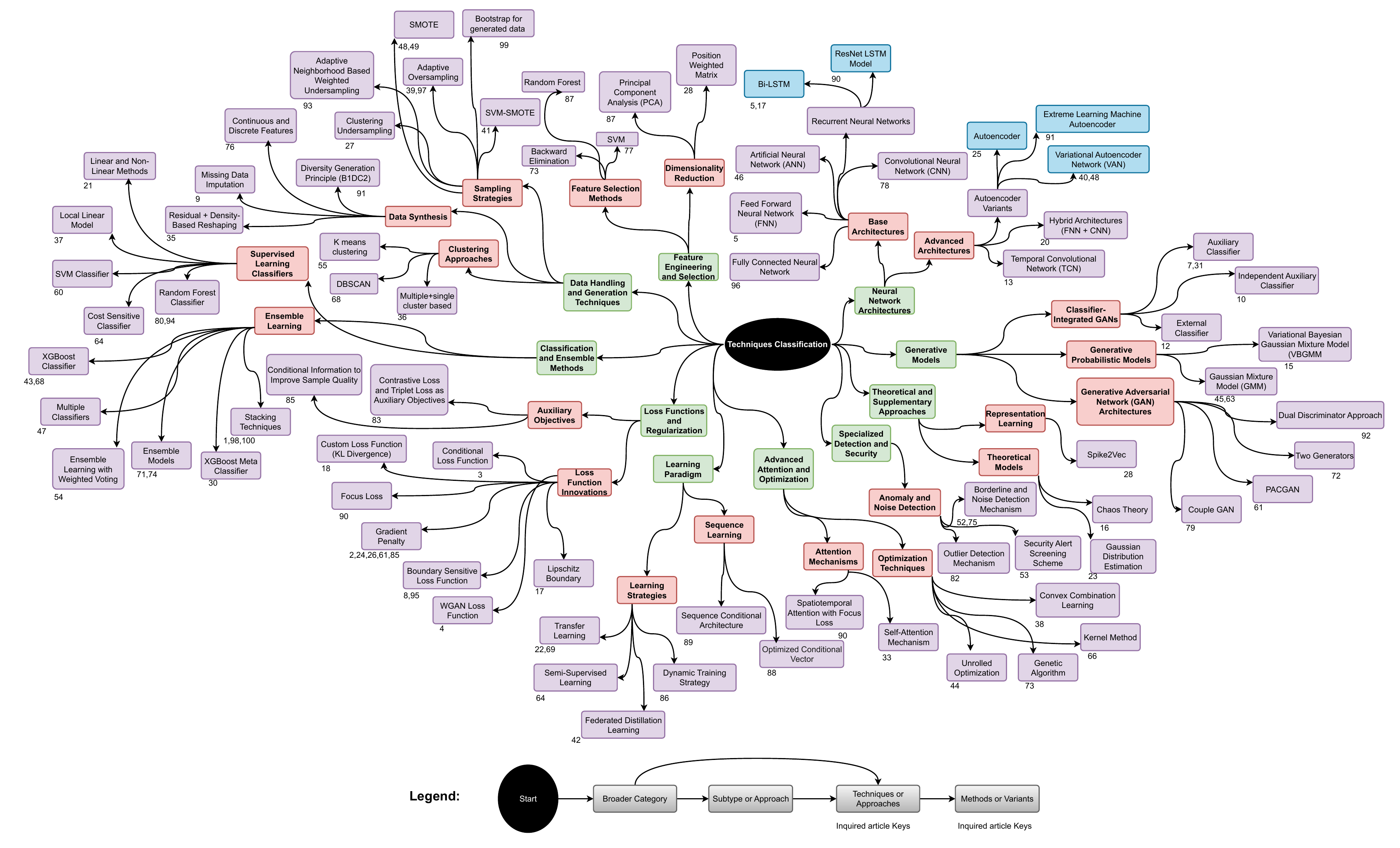}
    \caption{Categorization of techniques introduced or analyzed in the examined articles by their corresponding keys}
    \label{fig:pdf_figure}
\end{sidewaysfigure}
To complement these structural innovations, loss functions and regularization play a critical role in stabilizing GAN training and enhancing sample quality. Techniques such as WGAN loss, boundary-sensitive loss, and Lipschitz constraints help handle common issues like mode collapse and training instability. Additional objectives, such as contrastive loss, triplet loss, and gradient penalties, further refine the learning process by encouraging the generation of distinct and expected samples. \\
Alongside these internal refinements, data handling and generation techniques ensure that GAN-generated samples are diverse, representative, and well-suited for downstream tasks. SMOTE, clustering-based undersampling, and adaptive oversampling integrate seamlessly with GANs to balance class distributions. Other approaches, such as density-based reshaping and neighborhood-weighted sampling, improve minority class representation while reducing the risk of overfitting. \\
GANs are further strengthened by classification and ensemble methods, which use the synthetic data to improve predictive performance. Ensemble techniques, such as stacking, weighted voting, and multi-classifier systems, combine the strengths of algorithms like Random Forests, Support Vector Machines (SVMs), and XGBoost. These methods ensure robust decision-making when dealing with highly imbalanced datasets. \\
In addition to data generation, feature engineering and selection techniques optimize the preprocessing pipeline for GANs. Dimensionality reduction methods, such as Principal Component Analysis (PCA) and backward elimination, identify the most relevant features, while random forest-based selection methods improve the dataset's overall quality. These techniques ensure that GANs focus on generating and learning from the most critical aspects of the data. \\
Optimization techniques and attention mechanisms refine GAN performance by improving convergence and synthesis accuracy. Self-attention and spatiotemporal attention mechanisms enable GANs to focus on key features in structured or time-series datasets. Strategies like genetic algorithms and unrolled optimization fine-tune GAN parameters for efficient and stable training processes. \\

In specialized domains, detection and security use GANs' ability to identify rare events and anomalies. Techniques such as border and noise detection, outlier identification, and spatiotemporal attention are usually effective in intrusion detection, malware analysis, and other cybersecurity challenges. These mechanisms allow GANs to address imbalanced data in high-stakes environments. \\

Finally, theoretical and supplementary approaches contribute to a more in-depth understanding of GAN behavior and data aspects. Representation learning, Gaussian distribution estimation, and clustering methods such as K-means and DBSCAN are commonly used to analyze data distributions. Advanced theoretical concepts, including chaos theory, are occasionally utilized to improve GAN outputs in specific applications, broadening their applicability. \\
GANs have evolved into powerful tools for imbalanced data by using these diverse enhancements. From foundational improvements in architectures to advanced strategies in optimization, classification, and theoretical modeling, these techniques collectively show the adaptability and impact of GANs across a wide range of applications. 

\subsection{FI4:Which GAN variants are used for handling the imbalance issue?}

GANs have evolved into numerous variants to address the limitations of the original architecture and adapt to the complexities of imbalanced datasets. As highlighted in Figure~\ref{fig:png_figure}, these variants account for both traditional and novel approaches such as WGAN, CTGAN, CGAN, and WCGAN.

At the basis of this growth lies the Vanilla GAN, the original architecture that introduced adversarial training for synthetic data generation. Because of their simplicity, Vanilla GANs remain crucial to imbalanced data handling in smaller or less complex datasets. Among the reviewed works, 40 \% of all GAN variants outline their origins to Vanilla GAN, with 25 \%  representing newly developed modifications. These newer models address the challenges of mode collapse and unstable training by including advanced regularization and optimization techniques. While Vanilla GAN serves as a baseline, variants such as CycleGAN, GAME, GradGAN, OBGAN, and VAEGAN specialize in transformations between data distributions. These variants lead to creative ways the original framework has been adapted for modern challenges. These innovations highlight the enduring relevance of Vanilla GAN while expanding its applicability across diverse domains. \\

Building upon the limitations of Vanilla GAN, the Wasserstein GAN (WGAN) introduces the Wasserstein distance to improve training stability and the diversity of synthetic samples. WGAN and its derivatives constitute 15 \% of all GAN variants, with 13 \% comprising newly developed extensions that handle typical challenges in handling imbalance. Extensions such as WGAN-GP, which includes gradient penalties, and TWGAN-GP, which introduces additional regularization techniques, refine WGAN's ability to generate high-quality data. Additionally, WCGAN, a Wasserstein conditional extension, accounts for 2 \% of the total GAN variants, with an equal 2 \% representing new adaptations. They merge the stability of WGAN with class conditioning, allowing targeted data generation for minority classes. \\

Conditional GANs (CGANs) use a different process by allowing class-specific data handling. CGANs contribute to  20 \% of all GAN variants, with 16 \% representing new extensions. Their conditional mechanisms allow precise control over data synthesis, providing that generated samples align closely with the targeted class distributions. Variants such as ACTGAN and cWGAN combine auxiliary classifiers and conditional mechanisms. \\

Among the most widely used conditional variants is the CTGAN, which focuses on the tabular data and effectively handles mixed categorical and numerical features. CTGAN and its extensions constitute 23 \% of all GAN variants, with 12 \% representing new extensions. Unique adaptations such as K-Means CTGAN, which uses clustering techniques, and Copula-GAN, which uses probabilistic modeling. Domain-based extensions, including CTAB-GAN, optimized for structured tabular data, further present the flexibility of CTGAN in addressing unique challenges. \\

\begin{figure}[!htbp]
    \centering
    \includegraphics[width=\textwidth, height = 0.97\textheight]{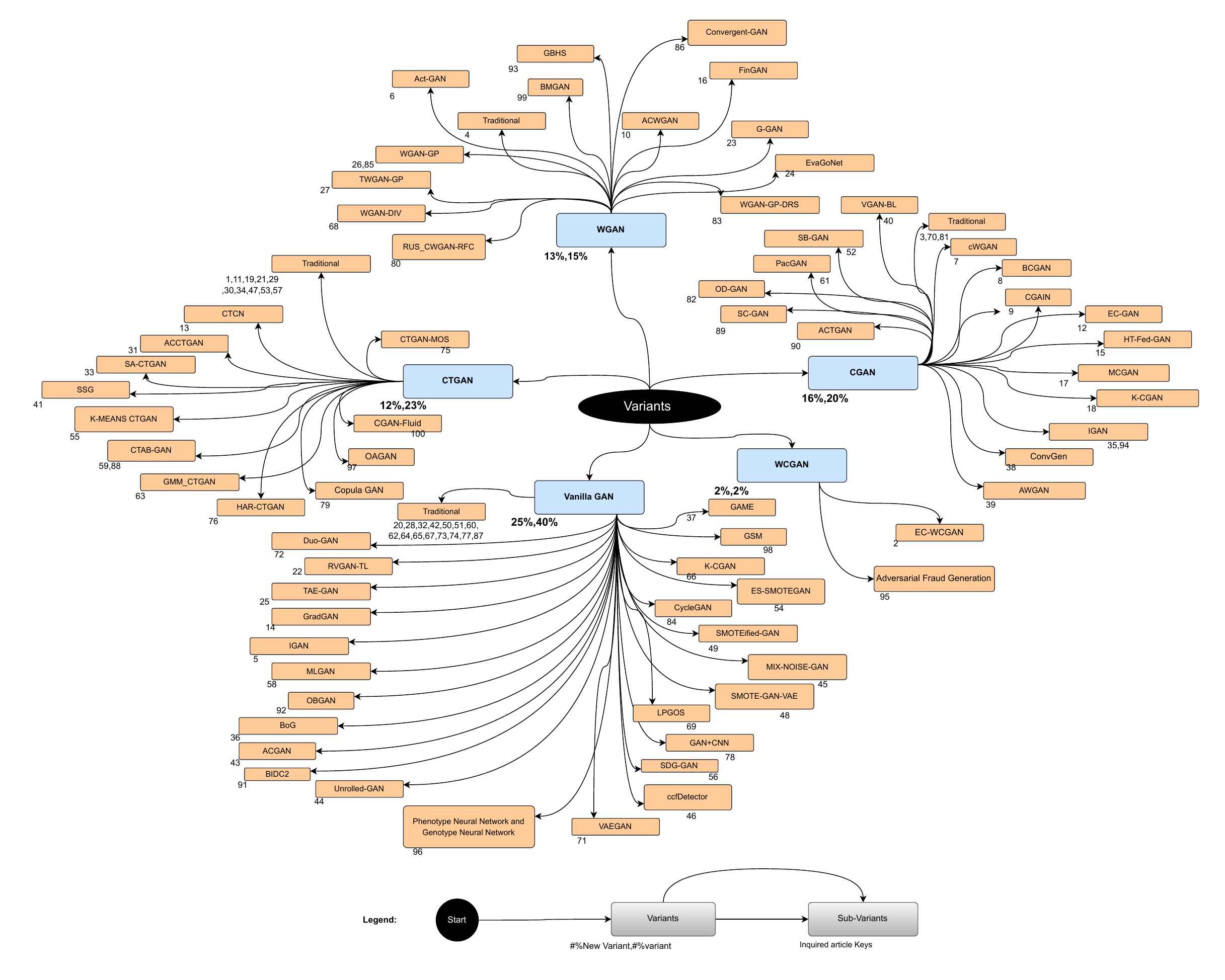}
    \caption{Categorization of variants introduced or analyzed in the examined articles by their corresponding keys}
    \label{fig:png_figure}
\end{figure}

 The evaluation of GAN variants presents a balance between preserving the foundational innovations of Vanilla GAN and exploring variants like WGAN, CGAN, and CTGAN to address the complexities of imbalanced problems. By using targeted extensions and domain-based insights, these variants have transformed GANs into robust tools for data augmentation, capable of meeting the diverse demands of modern applications.

\subsection{FI5: Which performance metric is used to compare GAN with traditional techniques?}
The evaluation of GANs with its traditional methods for handling imbalanced datasets also 
 considers dependency on a diverse range of performance metrics. Table~\ref{tab:metric_clusters} provides a comprehensive clustering-based approach of articles based on the metrics they use, showcasing how studies assess the usefulness of GANs across different scenerios. \\

Common classification metrics such as F1 Score, AUC ROC, accuracy, precision, and recall dominate compared to other metrics. These metrics are used in 90 articles, reflecting their widespread use due to their general applicability in classification tasks. The F1 Score is favored for imbalanced datasets as it balances precision and recall. At the same time, AUC ROC evaluates the trade-off between sensitivity and specificity, underling model performance across varying thresholds. \\

Imbalance-specific metrics include G-Mean, Sensitivity, Specificity, AUC PR (Area Under the Precision-Recall Curve), Balanced Accuracy (Bal-Acc), Ambiguity Score, and MAUC (Mean AUC), are reported in 34 articles. These metrics are required for datasets where class imbalance skews traditional performance measures. G-Mean balances sensitivity, while AUC PR focuses on evaluating models where false positives and false negatives carry unequal weights. Balanced accuracy is particularly useful for datasets with extreme imbalance. \\

\begin{table}[ht]
\centering
\caption{Clustering Articles by Metric Usage Patterns}
\label{tab:metric_clusters}
\begin{tabular}{|p{2cm}|p{4cm}|p{2cm}|p{2cm}|}
\hline
\textbf{Cluster} & \textbf{Metrics Used} & \textbf{Number of Articles} & \textbf{Article Keys} \\
\hline
\textbf{Common Metrics} & F1 Score, AUC ROC, Accuracy, Precision, Recall & \textbf{90 (overlapping)} & \textbf{1, 3, 4, 5, 6, 7, 10, 11, 12, 13, 14, 15, 16, 17, 18, 19, 20, 21, 22, 23, 24, 25, 26, 27, 28, 29, 30, 31, 32, 33, 35, 36, 39, 40, 41, 42, 43, 44, 45, 46, 47, 48, 50, 51, 52, 53, 54, 55, 56, 57, 58, 60, 61, 64, 65, 66, 67, 68, 69, 70, 71, 72, 73, 74, 75, 76, 77, 78, 79, 80, 81, 82, 83, 84, 85, 86, 87, 88, 89, 90, 91, 92, 93, 94, 95, 96, 97, 98, 99, 100
} \\
\hline
\textbf{Imbalance-Specific} & G-mean, Sensitivity, Specificity, AUC PR, MAUC (Mean Area Under Curve), Ambiguity Score, Bal-Acc (Balanced Accuracy) & \textbf{34 (overlapping)} & \textbf{2, 7, 10, 21, 23, 32, 35, 38, 39, 40, 44, 45, 46, 55, 57, 64, 65, 69, 71, 74, 75, 77, 81, 84, 85, 86, 91, 92, 93, 96, 97, 98, 99, 100
} \\
\hline
\textbf{Regression Metrics} &  RMSE, R2, MAE, MSE, MAPE, Brier Score & \textbf{8 (overlapping)} & \textbf{9, 20, 34, 59, 63, 64, 7, 62} \\
\hline
\textbf{Classification Metrics} & FPR, MCC, Balanced Error Rate, Kappa, TNR, PPV, NPV & \textbf{7 (overlapping)} & \textbf{2, 17, 39, 41, 53, 54, 100} \\
\hline
\textbf{Ranking Metrics} & DCG, Ranking Loss, Exact Match Ratio & \textbf{2 (overlapping)} & \textbf{58, 61 } \\
\hline
\textbf{Statistical Metrics} & KS Statistics, Silhouette Score & \textbf{2 (overlapping)} & \textbf{90, 91} \\
\hline
\textbf{Other Metrics} & Hamming Loss, MMD, Profit & \textbf{3 (overlapping)} & \textbf{58, 91, 93} \\

\hline
\end{tabular}

\end{table}

Regression-based metrics, including RMSE (Root Mean Square Error), MAE (Mean Absolute Error), MAPE (Mean Absolute Percentage Error), Brier Score, and R2 are used in 8 articles. These metrics are used in cases where GANs are used for generating continuous-valued outputs or improving regression models by handling imbalanced issues. \\

Classification metrics, including FPR (False Positive Rate), MCC (Matthews Correlation Coefficient), balanced error rate, Kappa, TNR (True Negative Rate), PPV (Positive Predicted Value), and NPV (Negative Prediction Value), are used in 7 articles. These metrics provide deeper understanding of the model's ability to handle imbalance. \\

In niche applications, ranking metrics like DCG (Discounted Cumulative Gain), ranking loss, and exact match ratio appear in 2 articles that mainly focus on tasks where ordering or prioritization of predictions is required. Similarly, statistical metrics such as KS Statistics and Silhouette Score are observed in 2 articles, which reflects their utility in analyzing the statistical alignment of GAN-generated data with the original datasets. 
Finally, a few articles report other metrics like Hamming Loss, MMD (Maximum Mean Discrepancy), and Profit, showing the growing demand for domain-based measures. \\

The clustering mapping of metrics shows the evaluation approaches used to compare GANs with traditional techniques. Meanwhile, standard metrics like F1 Score and AUC ROC remain dominant due to their familiarity and interpretability. Imbalance-specific metrics are crucial for capturing the performance in imbalanced scenarios. The inclusion of regression, statistical, and ranking metrics shows the expanding scope of GAN applications beyond standard classification tasks.

\subsection{TA1:How has research on GANs for imbalanced data evolved over time?}

The research on GANs for addressing imbalanced data has seen a significant growth trajectory since 2017, as illustrated in Figure~\ref{fig:png_figure}. Early adoption was minimal, with only one publication reported in 2017 and 2018. This initial phase reflects a period of conceptual exploration where foundational ideas on GANs for imbalanced data were introduced. \\

A notable increase began in 2019 with two publications, followed by steady growth in 2020, reaching eight publications. By 2021, the number of publications more than doubled to 22, marking the onset of the interest in applying GANs for data imbalance issues. The field peaked in 2023 with 42 publications, showing widespread recommendation of GANs as a critical tool in addressing class imbalance. However, a decline to 12 publications in 2024 indicates either a stabilization in the research trend or a shift in exploring other methodologies or domains. \\

The timeline presents an initial phase of conceptual foundation (2017-2018), a growth phase with increasing adoption (2019-2020), and a maturation phase with a peak in contributions (2022-2023), followed by a slight decrease in recent activity. \\ 

\begin{figure}[h!]
    \centering
    \includegraphics[width=\textwidth]{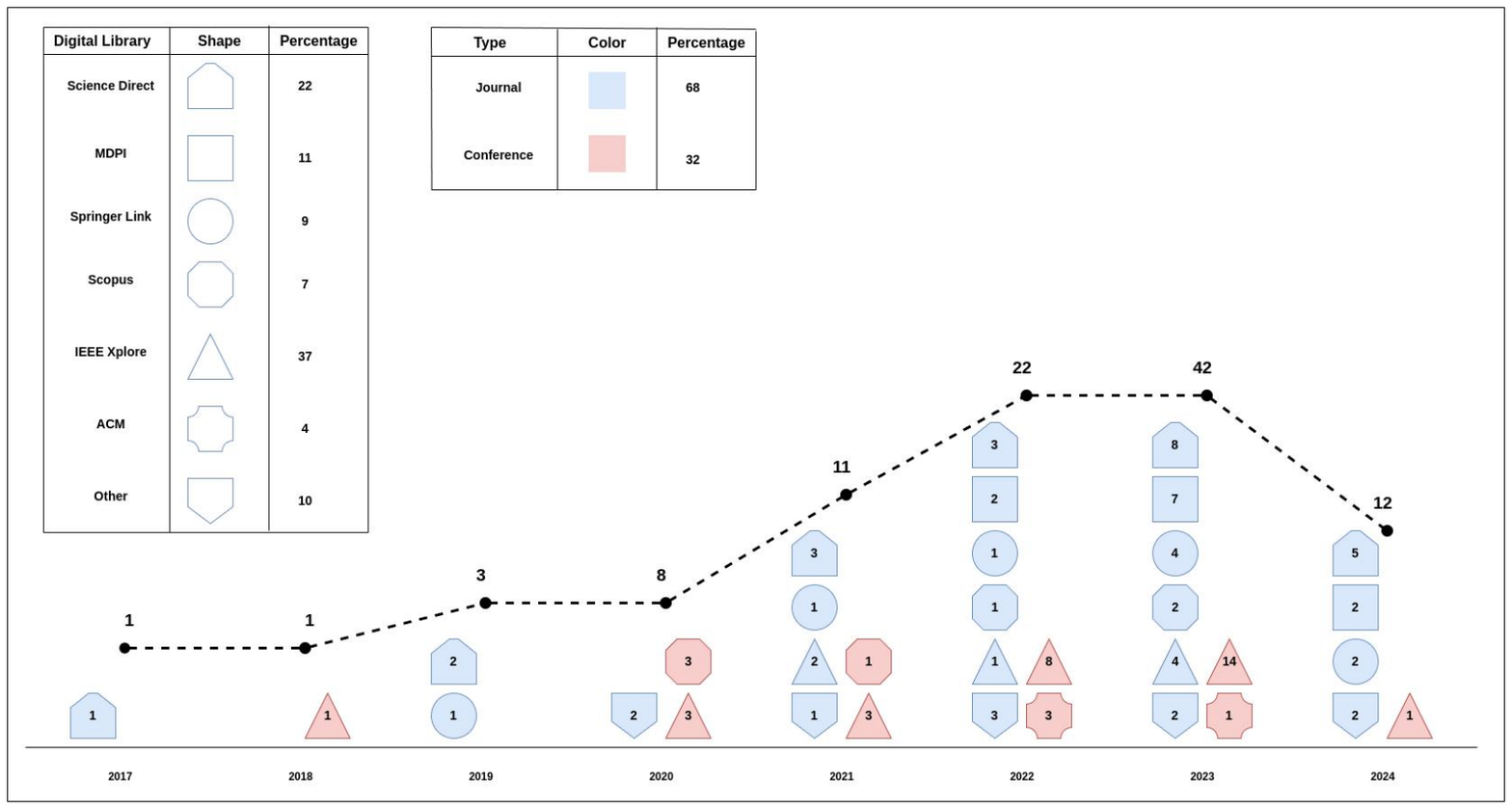}
    \caption{Categorization of studies and publications introduced or analyzed in the examined articles}
    \label{fig:png_figure}
\end{figure}

\subsection{TA2:Which journals, conferences, or platforms publish relevant work?}

The research on GANs for imbalanced data is distributed across various digital libraries, journals, and conferences, as shown in Figure~\ref{fig:png_figure}. Among these venues, IEEE Xplore leads with 37 \% of the total publications, indicating its usefulness in publishing cutting-edge research on GANs. ScienceDirect follows up with 22 \%, while MDPI and Springer Link include 11 \% and 9 \%, respectively. Platforms like Scopus (7 \%), ACM (4 \%), and other miscellaneous sources (10 \%) collectively include a smaller share of the research landscape. \\

In publication types, the majority of works (68 \%) are journal articles, presenting a deeper exploration. Conferences include 32 \% of the publications and act as a platform for presenting novel and preliminary findings. \\

The distribution presents the significant role of IEEE Xplore and ScienceDirect in shaping the GAN research landscape. At the same time, the mix of journals and conferences highlights a balance between theoretical and practical implementations. \\

\section{Learnings from the study}

Applying GANs to address imbalanced data has developed some critical observations, showing their broad adaptability across various domains and methodologies. The existing studies presented in the article have demonstrated several key learnings that support the progress and innovations in this field. These learnings explore domain-specific applications, advanced sampling techniques, architectural enhancements, performance evaluation metrics, research trends, and publication platforms. \\

Table~\ref{tab:Lessons Learned1} provides a comprehensive summary of these observations, supporting detailed information about the practical applications, methodological improvements, evaluation metrics, and transition channels that define GAN-based approaches to imbalance data challenges.\\

\begin{sidewaystable}
\centering
\caption{Learnings from the study}
\label{tab:Lessons Learned1}
\begin{tabularx}{\textwidth}{lX}
\hline\hline
\textbf{RI\#} & \textbf{Learnings from the study} \\ 
\hline\hline
\\
FI1 & GANs have shown remarkable versatility in addressing imbalanced data across various domains. They are broadly used in finance (35\%) for fraud detection, credit score, and cryptocurrency modeling, while mixed domain datasets (29\%) benefit from diverse data types. Healthcare (18\%) uses GANs for disease prediction and synthetic medical data generation; cybersecurity (12\%) applies them for intrusion detection and malware prevention; and other areas (6\%) explore unique applications like weather modeling and recommendation systems.
\\
\\
FI2 & GANs have gained importance in addressing class imbalance through data-level approaches. The focused approaches include oversampling and hybrid sampling techniques. GAN-based oversampling has been adopted in 83 studies with methods like density-based and clustering-focused techniques to further improve sample quality. Hybrid approaches adopted 17 studies and combined GANs with methods like SMOTE or undersampling to handle class overlap and noisy data. 
\\ 
\\
FI3 & GANs are increasingly augmented with algorithmic enhancements like exploring neural network architecture, generative models, loss functions, data handling methods, and classification techniques. Advanced architectures like CNNs, RNNs, and autoencoders promote high-quality sample synthesis, while other frameworks like PAC-GAN and dual-discriminator GANs improve density and class separation. Techniques like WGAN and contrastive loss stabilize training and improve synthetic data quality. Data handling methods like SMOTE and adaptive sampling improve class balance, and ensemble methods boost predictive performance. These enhancements, as a whole, show GANs adaptability and effectiveness in diverse applications. 
\\ 
\\
FI4 & GANs have evolved into diverse variants to address the challenges of imbalanced datasets, including Vanilla GAN, WGAN, CGAN, and CTGAN. Vanilla GAN is foundational to 40 \% of the variants and remains relevant for simpler datasets, while advanced adaptations like CycleGAN and GradGAN expand their utility. WGAN and its derivatives (15\%) improve stability and diversity with extensions like WGAN-GP and TWGAN-GP. CGANs (20\%) enable class-based generation with variants like ACTGAN adding auxiliary classifiers. CTGAN (23\%) excels in handling tabular data with specialized extensions like CTAB-GAN for structured datasets.
\\ 
\\

FI5 & The evaluation of GANs against traditional methods for handling imbalanced datasets uses various performance metrics. Classification metrics like F1 Score, AUC, ROC, accuracy, precision, recall are used in 90 articles. They dominate due to their general applicability, with F1 Score and AUC ROC used mainly for imbalanced data. Imbalanced-specific metrics such as G-Mean, AUC PR, and balanced accuracy were used in 34 articles that address skewed datasets. Regression metrics like RMSE and MAE (8 articles) and classification-specific metrics like MCC and Balanced Error Rate (7 articles) offer additional information. Niche metrics like DCG and statistical measures such as KS Statistics (2 articles each) show specialized applications. This diversity in metrics reflects the expanding coverage of GANs across classification, regression, and domain-based tasks.
\\ 
\\
TA1 & Research on GANs for addressing imbalanced data has grown significantly since 2017 and transitioned through distinct phases. The early conceptual phase (2017-2018) saw minimal adoption with just one publication each year. Research developments began in 2019 with two publications, followed by steady progress in 2020 (8 publications) and a surge in 2021 (22 publications), presenting growing mainstream interest. The field peaked in 2023 with 42 publications that justify GANs role in tacking class imbalance. A decline to 12 publications in 2024 suggests stabilization or a shift towards other approaches. This proves the evolution from foundation to maturity and potential transition.
\\
\\
TA2 & Research on GANs for imbalanced data is primarily transmitted through digital libraries and publishing platforms. IEEE Xplore dominates with 37\% of total publications, followed by ScienceDirect at 22\% and MDPI and Springer Link at 11\% and 9\% respectively. Other contributors include Scopus (7\%), ACM (4\%) and miscellaneous sources (10\%). Journals account for 68\% of publications with in-depth exploration, while conferences contribute 32\% showcasing innovative methods and preliminary findings. This distribution proves IEEE Xplore and ScienceDirect as key platforms for broadcasting GAN research, reflecting a balance between theoretical development and practical applications. 
\\ 
\\
\hline
\end{tabularx}
\end{sidewaystable}

\section{Conclusion}
This systematic mapping study reviewed the use of GANs in handling imbalanced data challenges in artificial intelligence. A ten-step filtering process led to the selection of 100 articles. The study mapped findings into domains of application, techniques, and GAN variants alongside the analysis of data level approaches and performance metrics, summarized in table~\ref{tab:Lessons Learned1}. \\

The findings show GANs broad applicability across fields, showing finance as the widely used domain, with GAN-based oversampling as a widely adopted and effective technique. Advanced architectures and tailored frameworks have further improved GAN outputs, handling challenges like class imbalance and noisy data. GAN variants like vanilla GANs, CTGANs and CGANs, are widely used in structured data handling, presenting the flexibility of GANs across diverse datasets and tasks. \\
Research interest in GANs for imbalanced data has grown steadily, peaking in recent years. Journals and conferences have complementary roles in offering theoretical and practical applications of GANs. However, the study's focus on tabular or structured data applications underlines a research gap in addressing other datasets like image. Hence, suggesting the future reviews include diverse criteria to include studies on well-known image and other datasets. \\
Finally, this mapping study presents the flexibility of GANs and their potential to address data imbalance. It also identifies opportunities for advancing hybrid approaches and using advanced neural network designs to improve their performance in artificial intelligence applications.

\backmatter

\bibstyle{sn-mathphys-num}
\bibliography{sn-bibliography}

\end{document}